\documentclass[journal]{new-aiaa}

\usepackage{natbib}
\usepackage{amsmath}
\usepackage[version=4]{mhchem}
\usepackage{siunitx}
\usepackage{longtable,tabularx}
\usepackage{amssymb}
\usepackage{graphics}
\usepackage{graphicx}
\usepackage{verbatim}
\usepackage{epsfig,float}
\usepackage{cite}
\usepackage{algorithm}
\usepackage{algorithmic}
\usepackage{amssymb}
\usepackage{url}
\usepackage{times,amsmath,epsfig}
\usepackage{tikz}
\usetikzlibrary{arrows}
\usepackage{verbatim}
\usepackage{float}
\usepackage{amsmath}
\usepackage{amssymb}
\usepackage{subfloat}
\usepackage{multirow}
\usepackage{graphicx}
\usepackage{subcaption}
\usepackage{color}

\DeclareMathOperator*{\argmax}{arg\,max}

\begin{document}

\title{A Deep Multi-Agent Reinforcement Learning Approach to Autonomous Separation Assurance}

\author{Marc W. Brittain \footnote{Ph.D. Candidate, Department of Aerospace Engineering, mwb@iastate.edu} and Xuxi Yang\footnote{Ph.D., Aerospace Engineering}}
\affil{Iowa State University, Ames, Iowa}
\author{Peng Wei\footnote{Assistant Professor, Department of Mechanical and Aerospace Engineering}}
\affil{George Washington University, Washington, DC}

\maketitle

\begin{abstract}
A novel deep multi-agent reinforcement learning framework is proposed to identify and resolve conflicts among a variable number of aircraft in a high-density, stochastic, and dynamic sector. Currently the sector capacity is constrained by human air traffic controller's cognitive limitation. We investigate the feasibility of a new concept (autonomous separation assurance) and a new approach to push the sector capacity above human cognitive limitation. We propose the concept of using distributed vehicle autonomy to ensure separation, instead of a centralized sector air traffic controller. Our proposed framework utilizes Proximal Policy Optimization (PPO) that we modify to incorporate an attention network. This allows the agents to have access to variable aircraft information in the sector in a scalable, efficient approach to achieve high traffic throughput under uncertainty. Agents are trained using a centralized learning, decentralized execution scheme where one neural network is learned and shared by all agents. The proposed framework is validated on three challenging case studies in the BlueSky air traffic control environment. Numerical results show the proposed framework significantly reduces offline training time, increases performance, and results in a more efficient policy.
\end{abstract}

%%%%%%%%%%%%%%%%%%%%%%%%%%%%%%%%%%%%%%%%%%%%%%%%%%%%%%%%%%%%%%%%%%%%%%%%%%%%%%%%
\section{INTRODUCTION}

\subsection{Motivation}
With the rapid expansion of global air traffic, guaranteeing aviation safety becomes a critical challenge. Air traffic is becoming more complex, not only in the structured airspace of commercial aviation, but also in low altitude airspace with small UASs and urban air mobility vehicles. Therefore, it requires certain level of autonomy to ensure safe separation in these environments. The tactical decisions being made today by human air traffic controllers have experienced little change in en route sectors over the past 50 years \citep{national2014autonomy}. The Advanced Airspace Concept (AAC), developed by Heinz Erzberger and his NASA colleagues, laid the foundation in autonomous air traffic control by developing tools such as the Autoresolver and TSAFE to augment human controllers for the increased airspace capacity and operation safety in conflict resolution 
\citep{erzberger2005automated,farley2007fast,erzberger2010algorithm}. Inspired by Erzberger, we believe that a highly automated ATC (ground-based) and/or separation assurance (onboard) system is the ultimate solution to support separation assurance of the high-density, complex, and dynamic air traffic in the future en route and terminal airspace.

Many recent proposals for low-altitude airspace operations, such as the UAS Traffic Management (UTM) \citep{kopardekar2016unmanned}, U-space \citep{undertaking2017u}, and urban air mobility \citep{mueller2017enabling}, require an autonomous air traffic control system or an autonomous separation assurance system to provide tactical de-confliction advisories or alerts to these intelligent aircraft, facilitate human operator decisions, and cope with high-density air traffic, while maintaining safety and efficiency \citep{mueller2017enabling,googleuas,air2015revising,kopardekar2015safely,balakrishnan2018blueprint,holden2016fast,uber}. According to a recent study by Hunter and Wei \citep{hunter2019service}, the key to these low-altitude airspace operations is to design the autonomous ATC system or separation assurance system on structured airspace to achieve the envisioned high traffic throughput. In this paper, we propose the concept of distributed separation assurance by individual aircraft, and focus on designing an onboard system that is able to provide real-time advisories to aircraft to ensure safe separation both along air routes and at air route intersections. Furthermore, we require this autonomous separation assurance system to be able to manage multiple intersections and handle uncertainty in a tactical manner.

Designing such an autonomous separation assurance system is challenging. We need a real-time onboard system to comprehend the current surrounding air traffic situation to provide advisories to aircraft in an efficient and scalable manner. One promising way to solve this problem is through reinforcement learning. The goal in reinforcement learning is to allow an agent to learn an optimal policy through interactions with an environment. The agent first perceives the state in the environment, selects an action based on the perceived state, and receives a reward based on this perceived state and action. By formulating the task of separation assurance as a reinforcement learning problem, we can obtain real-time advisories to aircraft with little computation overhead.

Artificial intelligence (AI) algorithms are achieving performance beyond humans in many real world applications today. AlphaStar, an AI agent built by DeepMind in 2019, is able to defeat the world's top professionals in StarCraft II, which is a highly complex and strategic game \citep{vinyals2017starcraft}. This notable advance in the AI field demonstrated the theoretical foundation and computational capability to potentially augment and facilitate human tasks with intelligent agents and AI technologies. However, such techniques require a fast-time simulator to allow the agent to interact with. The air traffic control simulator, BlueSky \citep{hoekstra2016bluesky}, developed by TU Delft allows for realistic real-time air traffic scenarios and we use this simulation software as the training environment and validation testbed of our proposed framework.

In this paper, a deep multi-agent reinforcement learning framework is proposed to enable autonomous air traffic separation in en route airspace, where each aircraft is represented by an agent. Each agent will comprehend the current air traffic situation and perform online sequential decision making to select speed advisories in real time to avoid conflicts, both at intersections and along route. Our framework is able to handle a variable number of aircraft in the sector and the complexity does not increase with the number of intersections. To achieve this we use attention networks \citep{bahdanau2014neural} to encode and control the importance of information about the environment (surrounding traffic scenario) into a fixed length vector (an abstract understanding). Our proposed framework provides another promising potential solution to enable an autonomous separation assurance system.

\subsection{Related Work}
There have been many important contributions to the topic of conflict resolution in air traffic control. One of the most promising and well-known lines of work is the Autoresolver designed and developed by Heinz Erzberger and his NASA colleagues \citep{erzberger2005automated,farley2007fast,erzberger2010algorithm}. It employs an iterative approach, sequentially computing and evaluating candidate trajectories, until a trajectory is found that satisfies all of the resolution conditions. The candidate trajectory is then output by the algorithm as the conflict resolution trajectory. The Autoresolver is a physics model-based approach that involves separate components of conflict detection and conflict resolution. It has been tested in various large-scale simulation scenarios with promising performance.

Strategies for managing intersections and metering fixes have been designed and implemented by NASA. These works include the Traffic Management Advisor (TMA) \citep{erzberger2014design} or Traffic Based Flow Management (TBFM), a central component of ATD-1  \citep{baxley2016air}. In this approach, a centralized planner determines conflict free time-slots for aircraft to ensure separation requirements are maintained at the metering fix. The main difference with our work is that we are dealing with multiple intersections instead of merging points so aircraft are required to maintain their current route and do not deviate at the intersections. The second difference is that our method is a decentralized framework that can handle uncertainty. In TMA or TBFM, once the arrival sequence is determined and aircraft are within the ``freeze horizon'' no deviation from the sequence is allowed, which could be problematic if one aircraft cannot meet the assigned time slot.

Conflict resolution techniques have also been investigated by multi-agent approaches in both structured and free airspace settings \citep{wollkind2004automated, yang2020scalable, bertram2020distributed, brittain2019autonomous}. In
\citep{wollkind2004automated}, negotiation techniques are introduced to resolve identified conflicts in the sector. In our research, we do not impose any negotiation techniques, but leave it to the agents to learn negotiation techniques through learning and training. \citep{yang2020scalable} introduces a message-passing based decentralized computational guidance algorithm using multi-agent Monte Carlo Tree Search (MCTS) formulation which is also able to prevent loss of separation (LOS) for UAS in an urban air mobility setting. \citep{bertram2020distributed} proposes a highly efficient Markov Decision Process (MDP) based decentralized algorithm that is also able to prevent LOS for cooperative and non-cooperative UAS in free airspace. In \citep{mollinga2020autonomous}, a graph neural network approach to conflict resolution in free airspace is proposed that is able to prevent collisions and conflicts in a scalabe manner. In previous work \citep{brittain2019autonomous}, we show that a decentralized formulation is able to resolve conflicts at an intersection, but this formulation only holds when the agent has access to the state information of the \textit{N}-closest agents, where the hyper-parameter \textit{N} needs to be selected and tuned through experimentation which limits the transferability of the network architecture to new environments. In our research, we allow the agent to have access to all aircraft state information and use attention networks to encode these information into a fixed length vector to handle a variable number of agents.

Challenging games such as Go, Atari, Warcraft, and most recently Starcraft II have been played by AI agents with beyond human-level performance \citep{vinyals2017starcraft,amato2010high,mnih2013playing,silver2016mastering}. These results show that a well-designed, sophisticated AI agent is capable of learning complex strategies under uncertainty. 

The first work in using reinforcement learning in a tactical air traffic control and conflict resolution application is described in \citep{brittain2018autonomous}. In this work, an AI agent is developed to mitigate conflicts, while also minimizing the delay of the aircraft in reaching their metering fixes. It was also shown later in \citep{pham2019machine} that an AI agent can resolve randomly generated conflict scenarios for a pair of aircraft by prescribing a vectoring maneuver for the ownship to implement. Both of these approaches fail to handle state space scalability as the number of intruder aircraft increase due to the single-agent architecture. In this work we are able to improve scalability through the concept of distributed separation assurance and alleviate the restriction to pair-wise conflict scenarios through the approach of multi-agent reinforcement learning. Our framework is built with a centralized learning architecture to encourage cooperative behavior among the aircraft to resolve any potential conflicts between themselves.

Recently the field of multi-agent collision avoidance has seen much success in using a decentralized framework in ground robots \citep{chen2017decentralized,everett2018motion}. In this work, the authors develop an extension to the policy-based learning algorithm (GA3C) that proves to be efficient in learning complex interactions between many agents. We find that the field of collision avoidance in robotics can be adapted to conflict resolution in aviation by considering larger separation distance.

Ground transportation research has already explored the use of deep reinforcement learning in the form of lane and speed change decision making \citep{wang2019lane,hoel2018automated}. These approaches place the agent on a single vehicle and treat surrounding vehicles as part of the environment. Our problem is similar to ground transportation in the sense we want to provide speed advisories to aircraft to ensure separation assurance, in the same way a vehicle wants to maintain a safe distance from surrounding vehicles. The main difference with our problem is that we are dealing with a multi-agent environment where each aircraft is represented by an agent. In this way we allow the agents to learn cooperative behavior through centralized learning and are able to provide separation assurance both along route and at the intersections.

In this paper, the deep multi-agent reinforcement learning framework is developed to solve the separation problem for variable number of aircraft for autonomous air traffic control in en route dynamic airspace where we avoid the computationally expensive forward integration method by learning a policy that can be quickly queried. The results show that our framework has very promising performance by producing a policy that maintains safe separation in an efficient manner.

The structure of this paper is as follows: in Section II, the background of reinforcement learning, policy based learning, multi-agent reinforcement learning, and attention network will be introduced. In Section III, the description of the problem and its mathematical formulation of deep multi-agent reinforcement learning are presented. Section IV presents our designed deep multi-agent reinforcement learning framework to solve this problem. The numerical experiments are presented in Section V. Section VI provides an analysis of the results and Section VII concludes this paper.

\section{Background}

\subsection{Reinforcement Learning}

Reinforcement learning, a branch of machine learning, is one type of sequential decision making where the objective is to learn a policy in a given environment with unknown dynamics. A reinforcement learning problem requires an interactive environment where an agent can select different actions that result in a change within the environment. If we let $t$ represent the current time step, then the components that make up a reinforcement learning problem are as follows:

\begin{description}
\item[$\bullet$ $S$  -] The state space $S$ is a set of all possible states in the environment
\item[$\bullet$ $A$  -] The action space $A$ is a set of all actions the agent can select in the environment
\item[$\bullet$ $r(s_{t},a_{t})$  -] The reward function determines how much reward the agent is able to acquire for a given ($s_{t}$, $a_{t}$) transition

\item[$\bullet$ $\gamma \in$] [0,1] - A discount factor determines how far in the future to look for rewards.  As $\gamma \rightarrow$ 0, immediate rewards are emphasized, whereas, when $\gamma \rightarrow$ 1, future rewards are prioritized.
\end{description}
$S$ contains all information about the environment and each element $s_{t}$ can be considered a snapshot of the environment at time $t$. Based on $s_{t}$, the agent is able to select an action $a_{t}$ which will affect the environment. The resulting change in the environment produces an updated state, $s_{t+1}$ and a reward associated from making the transition from $(s_t, a_t) \rightarrow s_{t+1}$. How the state evolves from $s_{t}$ $\rightarrow$ $s_{t+1}$ given action $a_{t}$ is dependent upon the dynamics of the environment, which is often unknown. In the reinforcement learning problem, the actions will be selected to maximize the cumulative reward, which needs to be carefully designed to reflect the objective of the environment. 

From this formulation, the agent is able to derive an optimal policy in the environment by maximizing a cumulative reward function. Let $\pi$ represent some policy and $T$ represent the total time for a given environment, then the optimal policy can be defined as follows:
\begin{equation}
    \pi^{*} = \argmax_{\pi}E[\sum_{t=0}^{T}(r(s_{t}, a_{t})|\pi)].
\end{equation}
By designing the reward function to reflect the objective in the environment, the optimal solution can be obtained by maximizing the total reward.

\subsection{Policy-Based Learning}
Reinforcement learning can be broken down into value-based and policy-based algorithms. 
% xuxi: replace following line
% In this work, we consider a policy-based reinforcement learning algorithm as these algorithms are able to learn stochastic policies, unlike value-based approaches. 
In this work, we consider a policy-based reinforcement learning algorithm where the basic idea is that a step in the policy gradient direction should increase the probability of better-than-average actions and decrease the probability of worse-than-average actions. Comparing with value-based algorithms, policy-based algorithms are able to learn stochastic policies.
This is especially beneficial in non-communicating multi-agent environments, where there is uncertainty in other agent's action. Proximal Policy Optimization (PPO) is a recent policy-based algorithm that uses a neural network to approximate both the policy (actor) and the value function (critic) \citep{schulman2017proximal}. PPO improved upon previous approaches such as A3C \citep{mnih2016asynchronous} by limiting the change from the previous policy to the new policy and has been shown to lead to better performance \citep{schulman2017proximal}. If we let $\theta_{\text{old}}$ and $\theta$ represent the neural network weights before and after the update at time $t$, $\zeta_{t}(\theta) = \pi_{\theta}(a_{t}|s_{t})/\pi_{\theta_{\text{old}}}(a_{t}|s_{t})$ is the term that describes the changing ratio of the policy after the update ($\zeta_t (\theta) = 1 $ means the policy does not change after the update). By restricting policy changing ratio in the range $[1-\epsilon, 1+\epsilon]$, the PPO loss function for the actor and critic can be formulated as follows:
\begin{equation}
\label{actor_loss}
    L_{\pi}(\theta) = 
    -E_{t}[\min(\zeta_{t}(\theta) \cdot A_t, \text{clip}(\zeta_{t}(\theta), 1-\epsilon, 1+\epsilon) \cdot A_t )] 
    - \beta\cdot H(\pi(s_{t}))
\end{equation}
\begin{equation}
\label{critic_loss}
    L_{v} = A_t^{2},
\end{equation}
where $\epsilon$ is a hyperparameter that bounds the policy changing ratio $\zeta_{t}(\theta)$. The second term $\beta\cdot H(\pi(s_{t}))$ in Equation~(\ref{actor_loss}) is used to encourage exploration by discouraging premature convergence to sub-optimal deterministic polices. Here $H$ is the entropy of the policy distribution and the hyperparameter $\beta$ controls the strength of the entropy regularization term.
In Equations~(\ref{actor_loss}) and (\ref{critic_loss}), the advantage function $A_t$ measures whether or not the action is better or worse than the policy's default behavior. In this paper we use the generalized advantage estimator GAE$(\gamma, \lambda)$ \citep{schulman2015high} to approximate $A_t$, which is defined as the exponentially-weighted average of the $k$-step advantage estimators:

% \begin{equation}
%     \hat{A}_{t}^{\mathrm{GAE}(\gamma, \lambda)}=(1-\lambda)(\hat{A}_{t}^{(1)}+\lambda \hat{A}_{t}^{(2)}+\lambda^{2} \hat{A}_{t}^{(3)}+\cdots)
% \end{equation}
\begin{equation}
    A_t=(1-\lambda)(\hat{A}_{t}^{(1)}+\lambda \hat{A}_{t}^{(2)}+\lambda^{2} \hat{A}_{t}^{(3)}+\cdots)
\end{equation}
where
\begin{equation}
    \begin{array}{ll}
    \hat{A}_{t}^{(1)}& =-V\left(s_{t}\right)+r_{t}+\gamma V\left(s_{t+1}\right) \\ 
    \hat{A}_{t}^{(2)}& =-V\left(s_{t}\right)+r_{t}+\gamma r_{t+1}+\gamma^{2} V\left(s_{t+2}\right) \\ \hat{A}_{t}^{(3)} & =-V\left(s_{t}\right)+r_{t}+\gamma r_{t+1}+\gamma^{2} r_{t+2}+\gamma^{3} V\left(s_{t+3}\right) \\ 
    &\cdots \\
    \hat{A}_{t}^{(k)} & =-V\left(s_{t}\right)+ \sum_{i=t}^{t+k-1}\gamma^{i-t} r_i +\gamma^{k} V\left(s_{t+k}\right) 
    % \hat{A}_{t}^{(k)} & =-V\left(s_{t}\right)+r_{t}+\gamma r_{t+1}+\cdots+\gamma^{k-1} r_{t+k-1}+\gamma^{k} V\left(s_{t+k}\right)
    \end{array}
\end{equation}

% In (3), the critic is trained to approximate the future discounted rewards, $R_{t} = \sum_{i=0}^{k-1}\gamma^{i}r_{t+i} + \gamma^{k}V(s_{t+k})$.

\subsection{Multi-Agent Reinforcement Learning}

\begin{figure}[t]
\begin{center}
\centerline{\includegraphics[width=0.7\columnwidth]{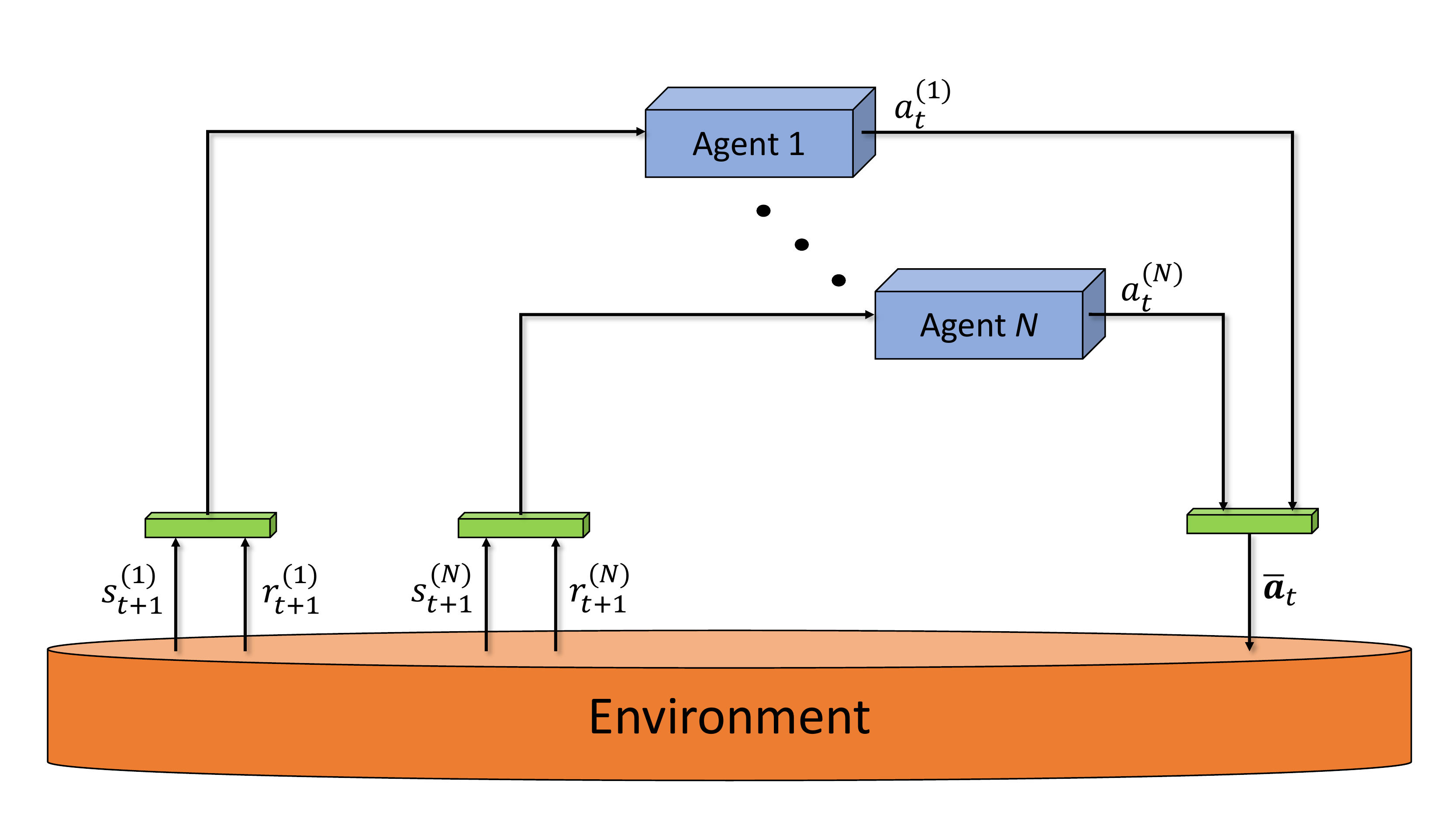}}
\caption{Progression of a multi-agent reinforcement learning problem. At time step $t$, each agent $i$ selects an action $a_t^{(i)}$ according to the learned policy. Then the environment will lead each agent $i$ to a new state $s_{t+1}^{(i)}$ with reward $r_{t+1}^{(i)}$.}
\label{marl}
\end{center}
\vskip -0.2in
\end{figure}

While single agent reinforcement learning considers one agent's interaction with an environment, multi-agent reinforcement learning is concerned with a set of agents that share the same environment \citep{bu2008comprehensive}. Figure~\ref{marl} shows the progression of a multi-agent reinforcement learning problem. Each agent has its own goals that it is trying to achieve in the environment that may be unknown to the other agents. The difficulty of learning useful policies greatly increases in these problems since the agents are both interacting with the environment and each other. One strategy for solving multi-agent environments is Independent Q-learning \citep{tan1993multi}, where other agents are considered to be part of the environment and there is no communication among agents. This approach often fails since each agent is operating in the environment and in return, results in learning instability. This learning instability is caused by the fact that each agent is changing its own policy and how the agent changes this policy will influence the policy of the other agents \citep{matignon2012independent}.

An alternative and popular approach to Independent Q-learning adopted in this paper is the centralized learning and decentralized execution where a group of agents can be trained simultaneously by applying a centralized method via an open communication channel \citep{kraemer2016multi}. Centralized learning that utilizes experiences from all of the agents can increase the learning efficiency. Decentralized policies where each agent can take actions based on its local observations have an advantage under partial observability and in limited communications during execution \citep{nguyen2018deep}. Centralized learning of decentralized policies has become a standard paradigm in multi-agent settings because the learning process may happen in a simulator where there are no communication constraints, and extra state information is available \citep{foerster2018counterfactual}.

\subsection{Attention}
One key limitation of many deep multi-agent reinforcement learning problems is that feedforward neural networks typically require a fixed-size input while the number of agents is dynamic. Comparing with general feedforward neural networks, recurrent neural networks (RNN) and most notably Long Short-Term Memory (LSTM) \citep{hochreiter1997long} networks are able to produce a fixed-size output from an arbitrary-length sequence state.
Although LSTMs are often applied to time sequences of data, we can leverage the LSTM's ability to encode a sequence of information that is not time-dependent by feeding each agent's state into the LSTM cell sequentially in descending order of their distances, to finally get a fixed-length vector representation of the variable number intruder aircraft information \citep{everett2018motion}.
However, the underlying assumption that the closest neighbors have the strongest influence is not always true \citep{vemula2018social}. Some other factors, such as speed and direction, are also essential for correctly estimating the importance of a neighboring agent, which reflects how this neighbor could potentially influence the ownship’s goal acquisition. In this work, we propose an attention mechanism for the agent to learn the relative importance of
each neighbor.

Recent work on attention networks have proven to achieve exceptional results when applied to neural machine translation tasks \citep{bahdanau2014neural} and multi-agent reinforcement learning problems \citep{baker2019emergent}. Unlike LSTM networks, attention networks have access to the global set of hidden vectors, whereas LSTM networks propagate the hidden vectors forward through time; potentially losing information from the past. 

If we now consider using attention (or LSTM) networks over agents instead of time, we obtain a network architecture that can handle a \textit{variable} number of agents. This is especially of interest as using a fixed number of agents could lead to important information not being included and result in sub-optimal policies.

With LSTM networks, how the agents are sorted before being processed is critical to ensure the most relevant information is retained. In \citep{brittainone}, agents were sorted based on the distance to the ownship. Distant aircraft were processed first, with closer aircraft processed last. While this approach is intuitive, there could be scenarios where more planning time is required to achieve the optimal policy, so some aircraft that are distant may be equally important to aircraft that are close. While the sorting strategy would need to be altered to account for this, attention networks have access to all aircraft and are not dependent on a sorting strategy. This allows the agent to \textit{learn} which aircraft are most important in selecting an optimal policy.

More specifically, the idea of an attentional model is to consider all the hidden states in the LSTM network when deriving the context vector $\boldsymbol{c}_{t}$. This allows the network to have global access to the sequence of hidden states to learn the importance of each hidden state $\overline{\boldsymbol{h}}_{s}$ with respect to the current target hidden state $\boldsymbol{h}_{t}$. In our work, we allow the attention network to directly operate on the variable length intruder information since we are using the attentional model over intruders instead of time. In general, a variable-length alignment vector ${\eta}(t, \cdot)$, whose size equals the number of elements in $\overline{\boldsymbol{h}}_{s}$, is derived by comparing the current target hidden state $\boldsymbol{h}_{t}$ with each hidden state in $\overline{\boldsymbol{h}}_{s}$:

% More specifically, the idea of an attentional model is to consider all the hidden states in the LSTM network when deriving the context vector $c_t$ that captures relevant source-side information to help the ownship make decision. In this model type, a variable-length alignment vector $\boldsymbol{a}_{t}$, whose size equals the number of time steps on the source side, is derived by comparing the current target hidden state $h_t$ with each source hidden state $\bar{h}_s$:
\begin{equation}
\label{bg_attn_weight}
\begin{aligned} \eta(t, s) &=\operatorname{align}\left(\boldsymbol{h}_{t}, \overline{\boldsymbol{h}}_{s}\right) \\ &=\frac{\exp \left(\operatorname{score}\left(\boldsymbol{h}_{t}, \overline{\boldsymbol{h}}_{s}\right)\right)}{\sum_{s^{\prime}} \exp \left(\operatorname{score}\left(\boldsymbol{h}_{t}, \overline{\boldsymbol{h}}_{s^{\prime}}\right)\right)} \end{aligned}
\end{equation}
where the score function is derived through the trainable parameter $\boldsymbol{W}_\eta$:

\begin{equation}
\label{bg_attn_score}
\operatorname{score}(\boldsymbol{h}_{t}, \overline{\boldsymbol{h}}_{s})=\boldsymbol{h}_{t}^{\top} \boldsymbol{W}_\eta \overline{\boldsymbol{h}}_{s}
\end{equation}

Given the alignment vector $\eta(t, s)$ as weights, the context vector $\boldsymbol{c}_{t}$ is computed as the weighted average over all elements of $\overline{\boldsymbol{h}}_{s}$:
\begin{equation}
\label{bg_attn_context}
\boldsymbol{c}_{t}=\sum_{s} \eta(t, s) \overline{\boldsymbol{h}}_{s}
\end{equation}

\section{Problem Formulation}

In en route and terminal sectors, air traffic controllers are responsible for ensuring safe separation among all aircraft. In our research, we used the BlueSky simulator as our air traffic control environment. We developed three challenging high-density air traffic case studies with varying number and orientations of routes with intersections to evaluate the performance of our deep multi-agent reinforcement learning framework.

\subsubsection{Objective} The objective in the case studies is to maintain safe separation (or prevent loss of separation) between aircraft and resolve conflicts for all aircraft in the sector by providing speed advisories. Unlike ground vehicles and UAS, commercial aircraft are unable to hover or stop at an intersection. This requires the agents to begin planning and coordinating with other agents well in advance of the intersections to ensure safe separation requirements. In order to obtain the optimal solution in this environment, the agents need to maintain safe separation and resolve conflicts at every time step in the environment. The designed case studies are dynamic, where aircraft enter the sector stochastically which provides a more difficult challenge for the agents since they need to develop a strategy instead of memorizing the reactive actions.
\subsubsection{Assumptions} There are several assumptions we made to make the case studies feasible. For each simulation run, there is a fixed maximum number of aircraft. This is to allow comparable performance between simulation runs and to evaluate the final performance of the model. In BlueSky, the performance metrics of each aircraft type impose different constraints on the range of cruise speeds. We set all aircraft to be the same type, Airbus A320, in the case studies. We also assume that all aircraft can not deviate from their route once determined. To achieve separation assurance, each aircraft has the ability to select between three different actions at each time step: maintain current speed, accelerate, and decelerate, with the minimum and maximum allowed speeds defined in BlueSky.

\subsection{Multi-Agent Reinforcement Learning Formulation}
Here we formulate our separation assurance case studies as a deep multi-agent reinforcement learning problem by representing each aircraft as an agent and defining the state space, action space, termination criteria, and reward function for the agents.
\subsubsection{State Space} The state contains the information an agent needs to make decisions. Since this is a multi-agent environment, we needed to incorporate communication and coordination between the agents. We assume that the position and dynamics of all intruders are available to each agent 
% (e.g. via automatic dependent surveillance—broadcast, or ADS–B).
(e.g. via automatic dependent surveillance-broadcast, or ADS-B).
The intruder's information is collected and sent through the attention network which produces a fixed length vector. This allows the agent to have access to all of the intruder's information without defining a maximum number to consider. In other words, this approach is able to handle variable number of agents in the environment. In previous work \citep{brittain2019autonomous}, only the \textit{N}-closest agents information was available to any given agent and in \citep{brittainone} the intruder aircraft were sorted based on distance through an LSTM. While these approaches work well, there are still limitations in generality: one still needs to tune the best value of \textit{N} for each environment and how to sort the intruders through the LSTM network. By using the attention network to encode the intruder information, we alleviate the limitations of the previous works by introducing a framework that can handle variable number of aircraft and is trained to know which intruder's information is most important for the ownship to make a decision, without pre-defining a sorting strategy.

The state information for the ownship includes the distance to the goal, aircraft speed, aircraft acceleration, a route identifier, and the loss of separation distance. Since we are dealing with structured airspace the position of each aircraft is represented as (distance to the goal, route identifier). The state information for each intruder includes the distance to the goal, aircraft speed, aircraft acceleration, a route identifier, distance from ownship to intruder, distance from ownship to intersection, and distance from intruder to intersection. The  intersection information is not included in the ownship state because for a given route, the ownship may encounter many intersections so the size of the state space would not scale with the number of intersections. An intersection can be defined as a potential conflict point between two aircraft so by leaving this information in the intruder state, we can handle pair-wise information in a scalable approach.

As described in \citep{brittain2019autonomous}, defining which agents to consider in the state space of the ownship is critical in obtaining optimal performance. We followed the same rules as in \citep{brittain2019autonomous} for defining which aircraft are allowed to be in the state of the ownship:

\begin{itemize}
    \item aircraft on conflicting route must have not reached the intersection;
    \item aircraft must either be on the same route or on a conflicting route.
\end{itemize}

More specifically, the state information for each agent is formulated as follows:

\begin{equation*}
    s^{o}_{t} = 
    (d_{\text{goal}}^{(o)}, v^{(o)}, a^{(o)}, r^{(o)}, \text{LOS}),
\end{equation*}
\begin{equation*}
    h^{o}_{t}(i) = 
    (d_{\text{goal}}^{(i)}, v^{(i)}, a^{(i)}, r^{(i)}, d_{o}^{(i)},d_{\text{int}}^{(o)},d_{\text{int}}^{(i)}),
\end{equation*}
where $s^{o}_{t}$ represents the ownship state information and $ h^{o}_{t}(i)$ represents state information for intruder $i$ that is available to the ownship.

\begin{figure*}[t]
\begin{subfigure}{0.50\textwidth}
  \centering
  \includegraphics[width=0.8\linewidth,height=4cm]{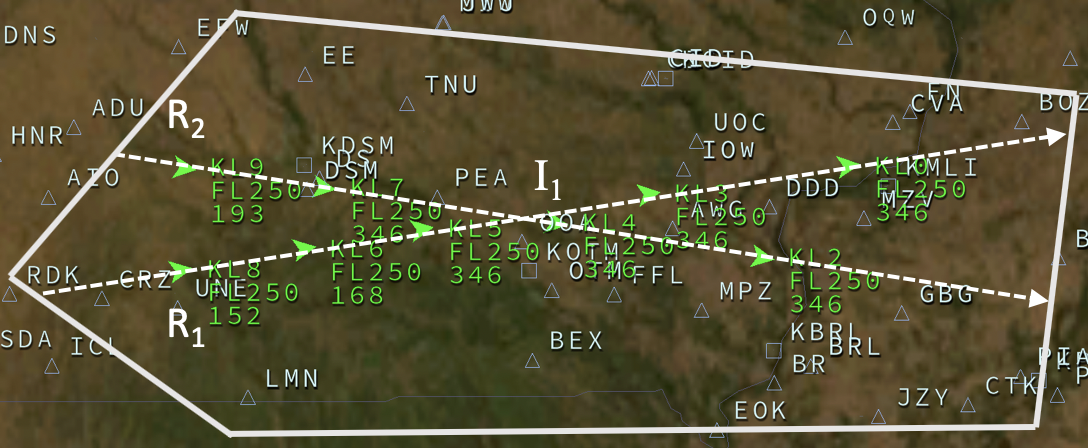}
  \caption{Case Study A}
  \label{fig:s1}
\end{subfigure}%\\
\begin{subfigure}{.50\textwidth}
  \centering
  \includegraphics[width=0.8\linewidth,height=4cm]{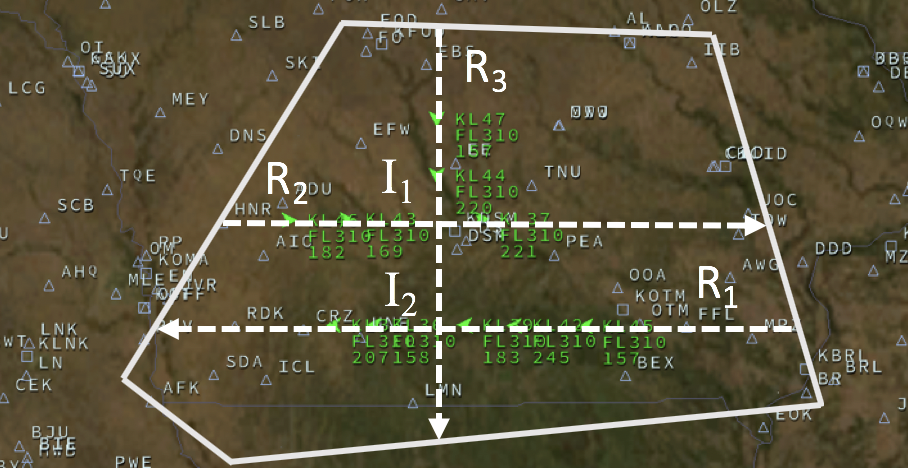}
  \caption{Case Study B}
  \label{fig:s2}
\end{subfigure}
\begin{center}
\begin{subfigure}{.50\textwidth}
  \centering
  \includegraphics[width=0.8\linewidth,height=4cm]{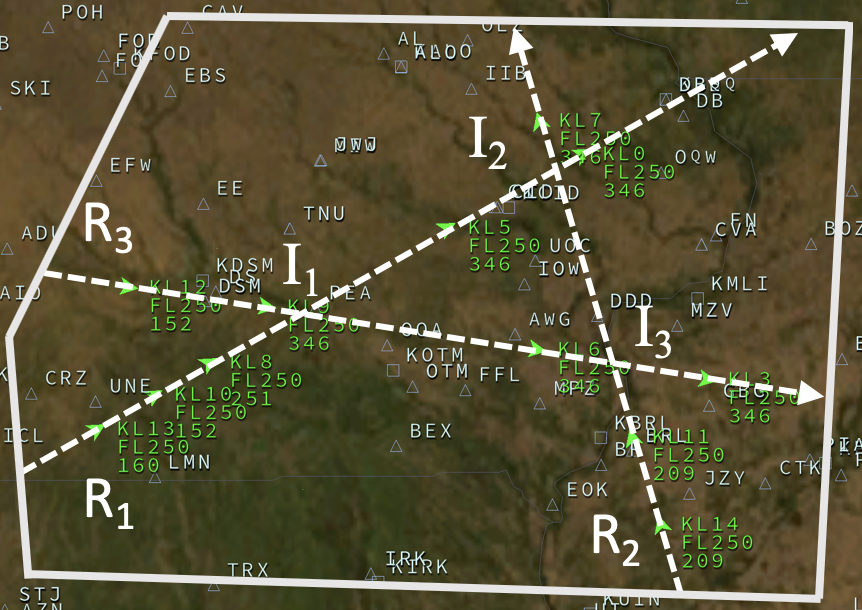}
  \caption{Case Study C}
  \label{fig:s3}
\end{subfigure}
\end{center}
\caption{Three Case Studies (three en route sectors) are designed in the BlueSky air traffic control environment to validate the performance of the proposed algorithm.}
\label{sector}
\end{figure*}

\subsubsection{Action Space}
To limit the number of actions sent to each aircraft, we allowed each agent to select an action every 12 seconds, which we also found to be sufficient to prevent loss of separation. The action space for the agents can be defined as follows:
\begin{equation*}
    A_{t} = [a_{-}, 0, a_{+}],
\end{equation*}
where $a_{-}$ is to decelerate (decrease speed), $0$ is no acceleration (hold current speed), and $a_{+}$ is acceleration (increase speed).

\subsubsection{Terminal State}
In each episode, a certain number of aircraft will be generated in the sector and this episode will terminate when all aircraft have exited the sector:
\begin{equation*}
    N_{\text{aircraft}} = 0.
\end{equation*}

\subsubsection{Reward Function}
Cooperation was encouraged by defining identical reward functions for all agents. The reward functions were locally applied so that if two agents were in conflict, they would both receive a penalty, but the remaining agents that were not in conflict would not receive a penalty. Here a conflict is defined as the distance between two aircraft is less than 3 nautical miles ($d^{\text{LOS}} =3$). The reward needed to be designed to reflect the goal of this paper: safe separation and conflict resolution. In previous work \citep{brittain2019autonomous,brittainone}, the designed reward function resulted in safe separation, but did not penalize speed changes. In a real world setting, speed changes should only be advised if a potential LOS event is to occur, otherwise the aircraft should maintain its current speed. We were able to capture our goals in the following reward function for the agents:

\begin{equation}
R(s,a) = R(s) + R(a),
\end{equation}

where $R(s)$ and $R(a)$ are defined as:

\begin{equation*}
  R(s) =
    \begin{cases}
      -1 & \text{if $d^{c}_{o} < d^{\text{LOS}}$}\\
      -\alpha + \delta\cdot d^{c}_{o} & \text{if $d^{\text{LOS}} \leq d^{c}_{o} < 10$} \\
      % -\alpha + \delta\cdot d^{c}_{o} & \text{if $d^{c}_{o} < 10$ and $d^{c}_{o} \geq d^{\text{LOS}}$}\\
       \;\;\, 0 & \text{otherwise}
    \end{cases}
\end{equation*}

\begin{equation*}
  R(a) =
    \begin{cases}
      \;\;\, 0 & \text{if $a$ = `Hold'}\\
      -\psi & \text{otherwise}
    \end{cases}.
\end{equation*}

In $R(s)$, $d^{c}_{o}$ is the distance from the ownship to the closest intruder aircraft in nautical miles, and $\alpha$ and $\delta$ are small, positive constants to penalize agents as they approach the loss of separation distance. $R(a)$ defines the reward penalty for speed changes. If $A_{t} = 0$ (hold), $R(a) = 0$, otherwise a penalty of $-\psi$ is introduced to minimize the number of speed changes. To ensure that the learned policy achieves safe separation while also minimizing speed changes, the magnitude of $\psi$ should be much smaller than $\alpha$. By defining the reward to reflect the distance to the closest aircraft, this allows the agent to learn to select actions to maintain safe separation requirements.

\section{Solution Approach}

We designed and developed a novel extension to our \textit{Deep Distributed Multi-Agent Variable} (D2MAV) framework \citep{brittainone} that incorporates (1) attention networks instead of LSTM networks, (2) a redefined reward function to minimize speed changes, and (3) a parallelized implementation to reduce wall-clock training time. We refer to this framework as D2MAV-A. In this section, we introduce and describe the framework, then we explain why this framework is needed to solve the case studies.

\begin{figure}
\vskip -0.2cm
\begin{center}
\centerline{\includegraphics[width=0.75\columnwidth]{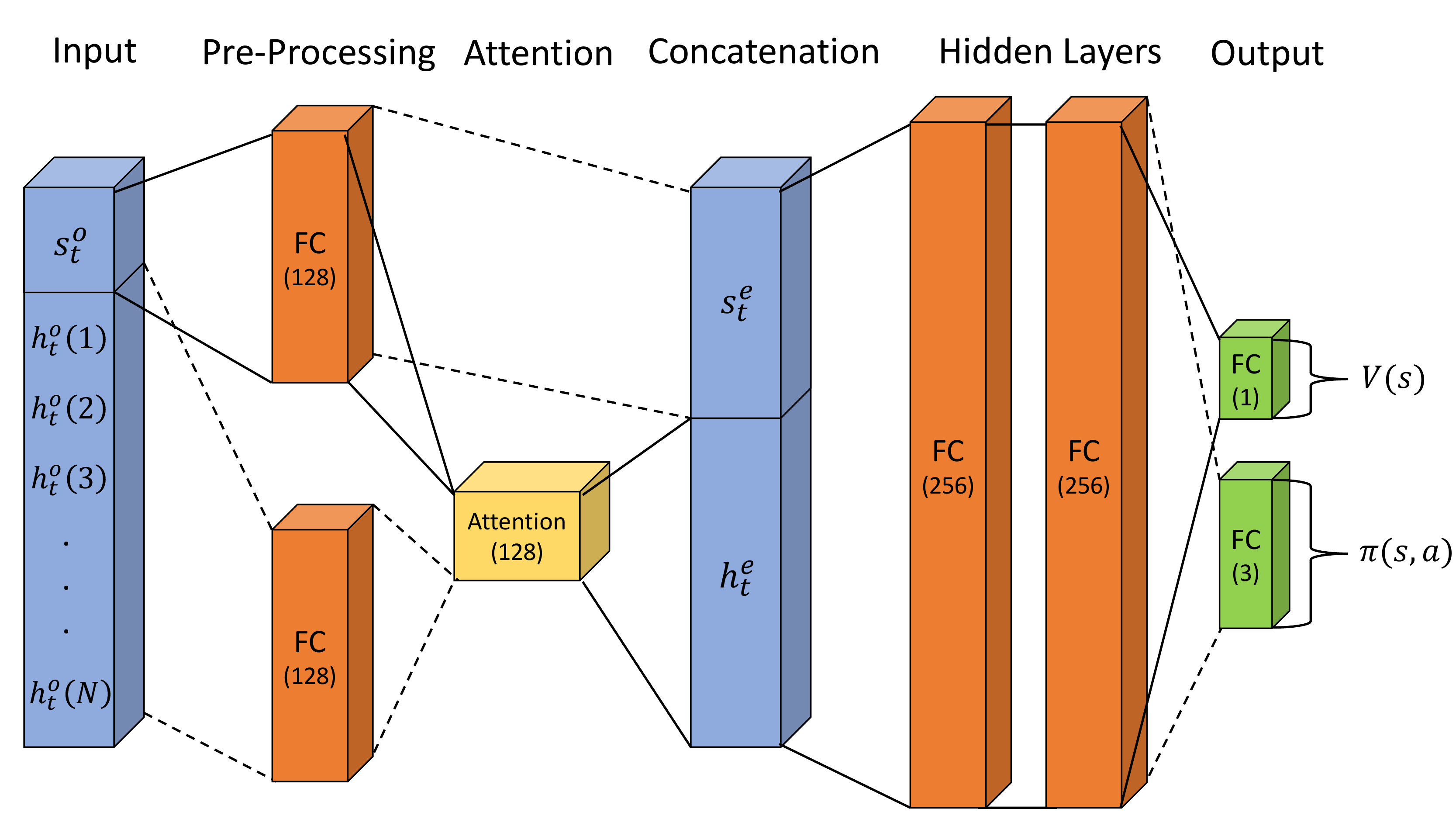}}
\caption{Illustration of the neural network architecture for the PPO algorithm with shared layers between the actor and critic. The intruder information is first pre-processed through a fully connected layer and an attention network to encode the intruder information to a fixed length vector. The fixed length vector is then concatenated with the ownship state information before being sent through two fully connected layers (FC) of 256 nodes. The policy and value is then obtained from the output fully connected layer of 4 nodes.}
\label{nn}
\end{center}
\vskip -0.4cm
\end{figure}

To formulate this environment as a deep multi-agent reinforcement learning problem, we utilized a centralized learning with decentralized execution framework with one neural network where the actor and critic share layers of the same neural network, further reducing the number of trainable parameters. By using one neural network that is shared by all agents, we can train a model that improves the joint expected return of all agents in the sector, which encourages cooperation. Our reinforcement learning algorithm is a policy-based approach, PPO, which is shown to perform well across a range of challenging environments \citep{schulman2017proximal}. We first pre-process the ownship and intruder state through fully connected layers to ensure consistent feature dimensions. We then use an attention network to encode the intruder state information to a fixed length vector. If we let $s$ represent the pre-processed ownship state and $\bar{h}_i$ represent the pre-processed state of intruder $i$ ($i \in \{1, \cdots, n\}$ where $n$ is total number of intruder aircraft that keeps changing in the simulator), we can define the attention network as follows:
\begin{equation}
    \label{eq:attention1}
    \text{score}(s,\bar{h}_i) = s^{\top} W_{1} \bar{h}_i\\
\end{equation}
\begin{equation}
    \label{eq:attention2}
    \eta_{s,\bar{h}_i} = \frac{\text{exp}(\text{score}(s,\bar{h}_i))}{\sum_{j=1}^{n}\text{exp}(\text{score}(s,\bar{h}_j))}
\end{equation}
\begin{equation}
    \label{eq:attention3}
    c_{s} = \sum_{i=1}^n\eta_{s,\bar{h}_i}\bar{h}_i
\end{equation}
\begin{equation}
    \label{eq:attention4}
    a_{s} = f(c_{s}) = \text{tanh}(W_{2}c_{s})
\end{equation}
where $\eta_{s,\cdot}$ is the attention weights of the ownship with respect to all of the other intruder aircraft, $c_{s}$ is the context vector that represents the weighted contribution of the surrounding air traffic, and $a_{s}$ is the attention vector that represents the abstract understanding of the surrounding air traffic. To calculate the score we use Luong's multiplicative style \citep{luong2015effective}. In our formulation we slightly modified Equation~(\ref{eq:attention4}) from the original formulation to concatenate the ownship state with the attention vector $a_s$ instead of the context vector $c_s$. This way, the agent now has access to the full ownship state, along with the attention vector. From there, the combined state is sent through two fully connected layers and produces two outputs: the policy and value for a given state. Figure~\ref{nn} shows an illustration of the neural network architecture. 

With this framework, we can implement the neural network's policy to all aircraft at the beginning of each episode. Each aircraft then follows this policy until termination. Since the environment is stochastic, it is difficult to evaluate the performance of the policy on one episode as the policy may perform well on one episode and poorly on another. Therefore, by collecting multiple episodes worth of experiences (state, action, reward, terminal) to update the neural network policy, the network can observe different outcomes from the same policy. In previous work \citep{brittain2019autonomous, brittainone}, five episodes were sequentially collected to update the neural network policy. In the D2MAV-A framework we leverage new parallel processing techniques in Python to collect 30 episodes of BlueSky in parallel to update the neural network policy, leading to better performance in substantially less time.

\subsection{Baselines}
As mentioned earlier, how the agents are sorted before processing through the LSTM determines how much information is retained (i.e. agents processed last contribute most to the encoded state). In \citep{brittainone} agents are sorted based on the distance to ownship and in \citep{brittain2019autonomous}, only the \textit{N}-closest agents are added to the state of the ownship.

\begin{figure}[t]
\begin{center}
\centerline{\includegraphics[width=0.5\columnwidth]{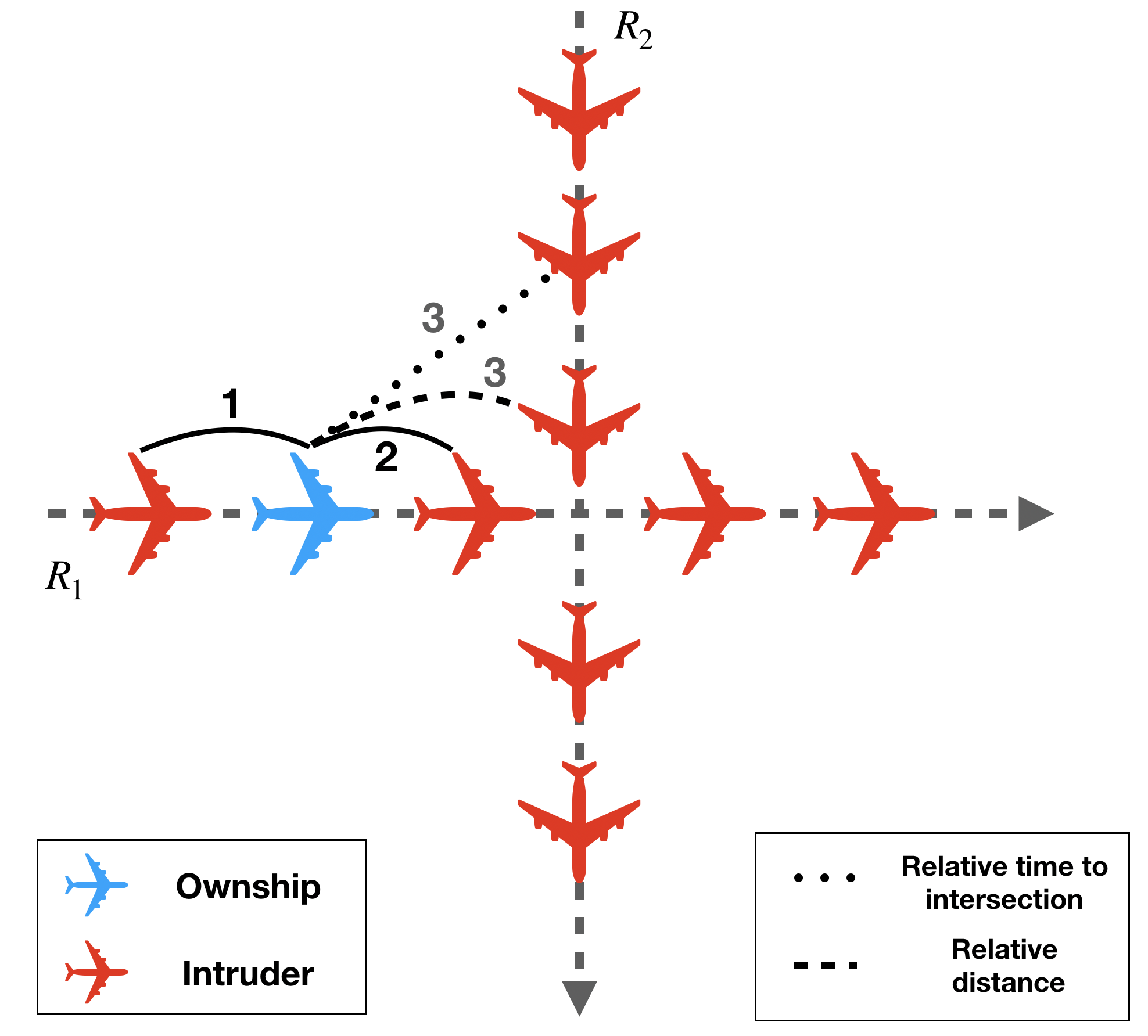}}
\caption{Illustration of the sorting strategies used in the baselines (for clarity, we only show three intruder aircraft considered for the ownship). If using the relative distance from the ownship, the three intruders that will be included for the ownship will be 1, 2, and the dashed 3. When using the relative time to the intersection, the intruders included for the ownship will be 1, 2, and the dotted 3. Although the dashed 3 intruder is \textit{closer} to the ownship, this intruder will pass the intersection before the ownship arrives and is not as relevant as the dotted 3 intruder that is closer in time.}
\label{sorting_strategy}
\end{center}
\vskip -0.2in
\end{figure}

To provide a fair comparison of the D2MAV-A framework to the D2MAV and D2MA frameworks, we introduce an additional sorting strategy that is based off of the relative time to the intersection for both D2MAV and D2MA, which we refer to as D2MAV-Time and D2MA-Time. In our evaluation of the D2MA framework we set the number of closest intruders, \textit{N} = 5.  Figure~\ref{sorting_strategy} provides an illustration of this sorting methodology. We also provide an additional random baseline where every action is selected uniformly at random from the action space to show the minimum performance in the case studies.

\section{Numerical Experiments}

\subsection{Simulation Environment}
We used BlueSky air traffic control simulator to evaluate the performance of the D2MAV-A framework along with our baselines. By design, when restarting the simulation, all objectives were the same: maintain safe separation and sequencing and resolve any potential conflicts. Aircraft initial positions and available speed changes did not change between simulation runs.

\subsubsection{Parallel BlueSky}
Sometimes approaches in academia fail to transfer to real world environments due to the conditions that the approaches were developed in. In deep reinforcement learning, most testing environments are designed to be massively parallelized such that algorithms can benefit from the increase in computation and data. BlueSky is an example of a real world simulator that was not designed for reinforcement learning but can be used to test reinforcement learning algorithms on. Since BlueSky was not designed for environment parallelization, previous work \citep{brittain2019autonomous, brittainone} relied on single-threaded processing. In this work, we leverage a new parallel processing module in Python, Ray \citep{moritz2018ray}, to create a parallelized extension to the BlueSky simulator that reinforcement learning algorithms can leverage for multi-thread use. This allows for multiple workers to all access BlueSky independently in their own thread. Each worker is then able to send their simulation results to the main scheduler worker to update the global model parameters. This extension will greatly advance the use and research of reinforcement learning algorithms in air traffic management. Our parallelized implementation of BlueSky will be available at https://github.com/marcbrittain. 

\subsection{Environment Setting}
Each simulation run in the BlueSky environment is referred to as an episode. An episode begins with one aircraft on each route and concludes when all aircraft have reached their terminal state. 

There were many different parameters that needed to be tuned and selected. We implemented the PPO concept mentioned earlier with a pre-processing fully connected layer for both the ownship and intruders that consisted of 128 nodes. We then process this information through an attention network with 128 nodes, before concatenating the fixed length intruder vector with the pre-processed ownship vector. We then feed the concatenated vector through two hidden layers consisting of 256 nodes. We used the Leaky ReLU activation function \citep{maas2013rectifier} for all hidden layers except the attention network for which we used a linear activation function. The output of the actor used a softmax activation function and the output of the critic used a linear activation function. We also used the Adam optimizer \citep{kingma2014adam} for both the actor and critic loss. Other key parameter values are shown in Table~\ref{param}.

\begin{table}[hb]
\caption{Finalized hyperparameters for D2MAV-A.}
\label{param}
\begin{center}
\begin{small}
\begin{sc}
\begin{tabular}{lcccr}
\hline
Parameter & Value \\
\hline
Learning Rate    & 0.0001   \\
GAE Discount Factor $\gamma$ & 0.99 \\
GAE Discount Factor $\lambda$ & 0.95 \\
PPO Ratio Bound $\epsilon$ & 0.2 \\
Reward Coefficient $\alpha$ & 0.1  \\
Reward Coefficient $\delta$ & 0.05 \\
Reward Coefficient $\psi$ & 0.001 \\
Entropy Coefficient $\beta$  & 0.0001 \\
Parallel Workers & 30 \\
Leaky Relu Alpha & 0.2 \\
\hline
\end{tabular}
\end{sc}
\end{small}
\end{center}
\end{table}

\begin{table*}[t]
\centering
\caption{Performance of the policy tested for 200 independent episodes.}
\resizebox{14.5cm}{!}{%
\begin{tabular}{ccccccc}
                                    & \multicolumn{6}{c}{Case Study}                                                                                                                                    \\ \cline{2-7} 
\multicolumn{1}{c|}{}               & \multicolumn{2}{c|}{A}                               & \multicolumn{2}{c|}{B}                               & \multicolumn{2}{c|}{C}                              \\ \cline{2-7} 
\multicolumn{1}{c|}{Framework}      & Mean                   & \multicolumn{1}{c|}{Median} & Mean                   & \multicolumn{1}{c|}{Median} & Mean                  & \multicolumn{1}{c|}{Median} \\ \hline
\multicolumn{1}{c|}{\textbf{D2MAV-A}}      & \textbf{29.99 $\pm$ 0.141}          & \multicolumn{1}{c|}{30}     & \textbf{30 $\pm$ 0.0}         & \multicolumn{1}{c|}{30}     & \textbf{30.0 $\pm$ 0.0} & \multicolumn{1}{c|}{30}     \\
\multicolumn{1}{c|}{D2MAV}           & 29.95 $\pm$ 0.312                      & \multicolumn{1}{c|}{30}      & 29.61 $\pm$ 0.865          & \multicolumn{1}{c|}{30}     & 29.57 $\pm$ 0.997                    & \multicolumn{1}{c|}{30}      \\
\multicolumn{1}{c|}{D2MA}      & 29.88 $\pm$ 0.475          & \multicolumn{1}{c|}{30}     & 29.93 $\pm$ 0.368          & \multicolumn{1}{c|}{30}     & 28.76 $\pm$ 1.56          & \multicolumn{1}{c|}{30}     \\
\multicolumn{1}{c|}{D2MAV-Time}      & 29.97 $\pm$ 0.243 & \multicolumn{1}{c|}{30}     & 29.32 $\pm$ 1.21           & \multicolumn{1}{c|}{30}     & 29.36 $\pm$ 1.18          & \multicolumn{1}{c|}{30}     \\
\multicolumn{1}{c|}{D2MA-Time} & 29.84 $\pm$ 0.578          & \multicolumn{1}{c|}{30}     & 29.94 $\pm$ 0.395 & \multicolumn{1}{c|}{30}     & 28.52 $\pm$ 1.70          & \multicolumn{1}{c|}{28}     \\
\multicolumn{1}{c|}{Random}         & 17.77 $\pm$ 2.86           & \multicolumn{1}{c|}{18}     & 10.36 $\pm$ 3.10           & \multicolumn{1}{c|}{10}     & 17.45 $\pm$ 2.80          & \multicolumn{1}{c|}{18}     \\ \hline
\end{tabular}
}
\label{policy_perf}
\end{table*}
\begin{figure}[t]
\begin{center}
\vskip -0.2cm
\centerline{\includegraphics[width=0.75\columnwidth]{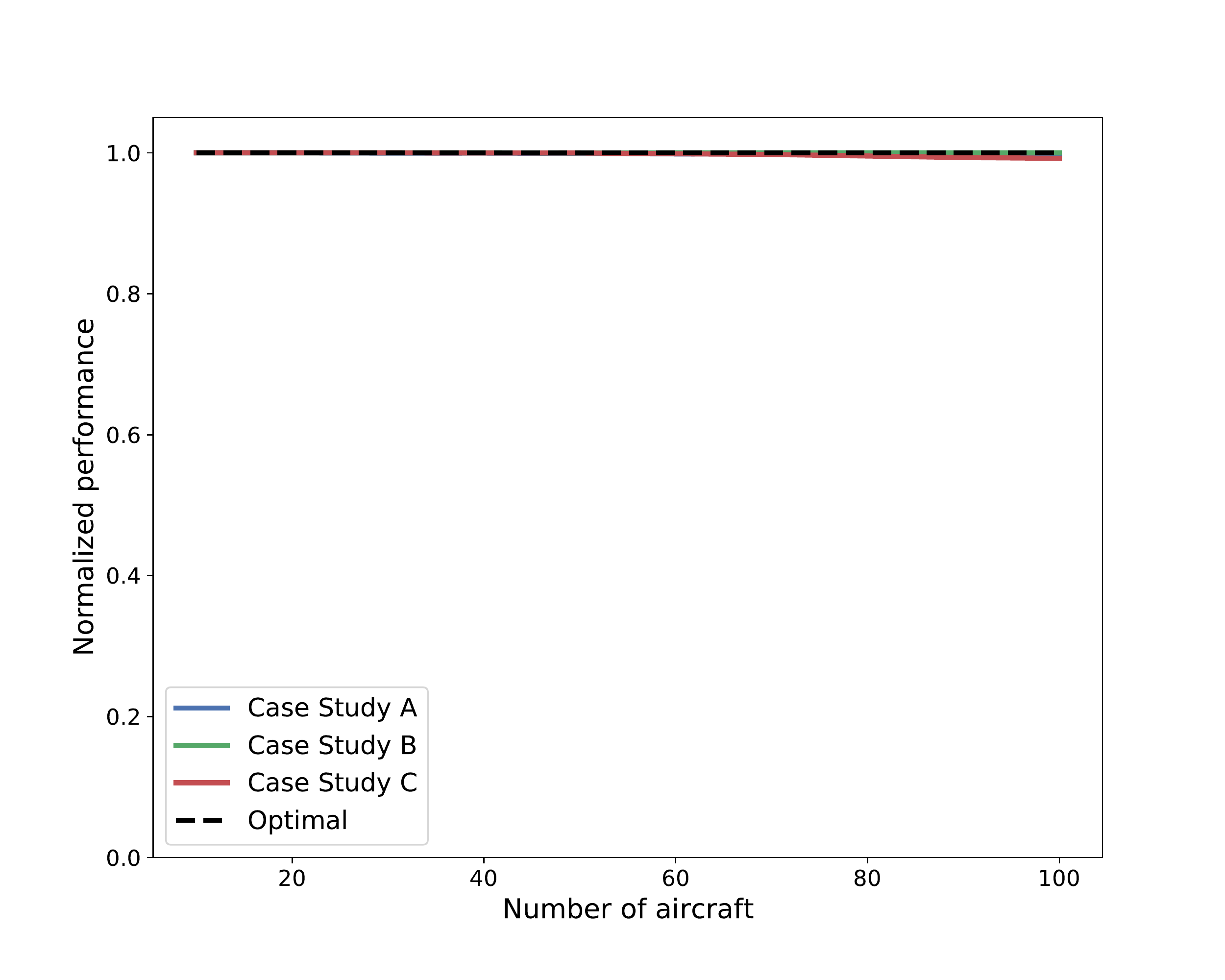}}
\caption{Normalized performance of the D2MAV-A framework on each of the case studies. A score of 1 represents optimal performance.}
\label{norm_perf}
\end{center}
\vskip -0.2cm
\end{figure}
\begin{figure*}[tb]
\begin{subfigure}{0.50\textwidth}
  \centering
  \includegraphics[width=\linewidth]{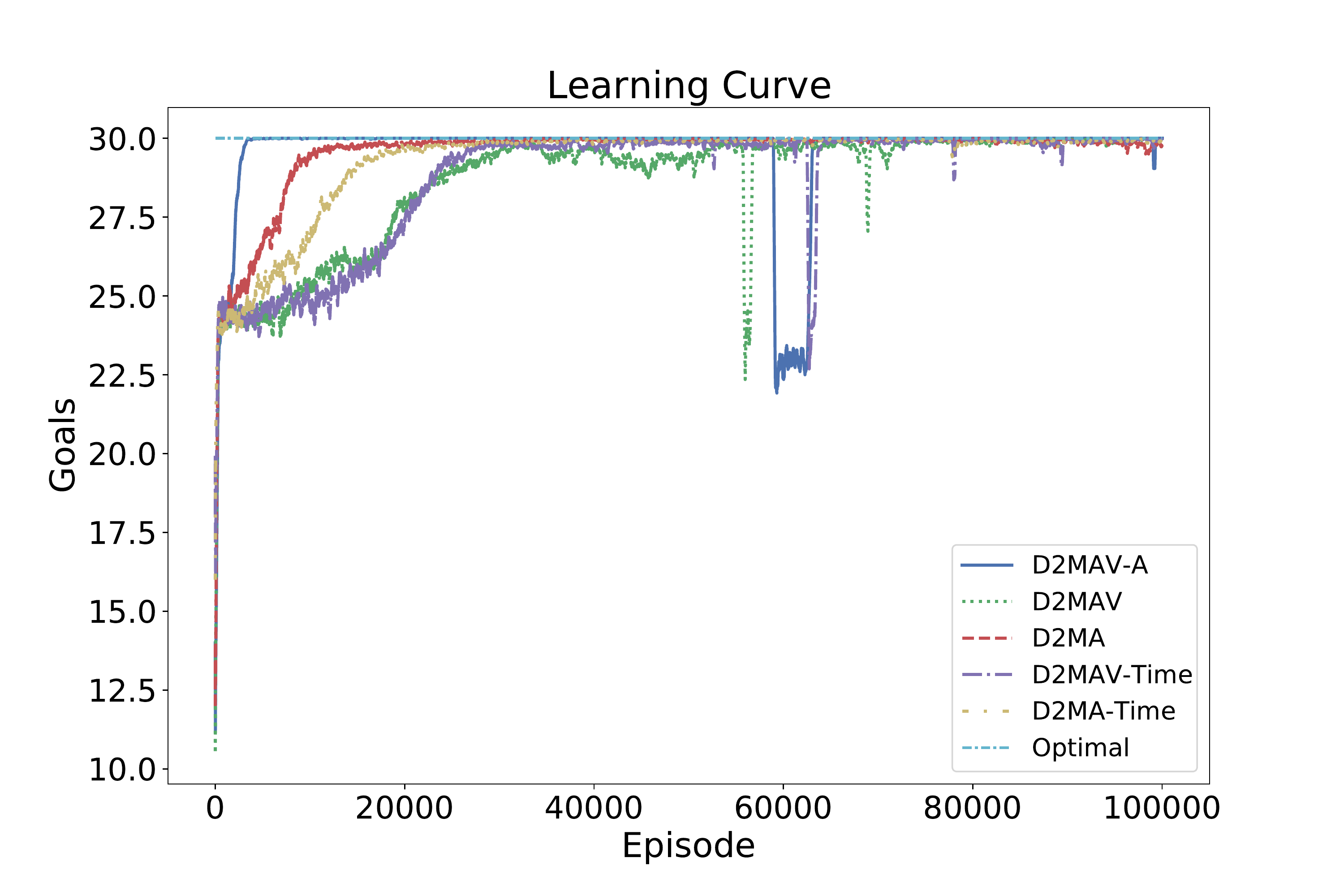}
  \caption{Case Study A}
  \label{fig:lc1}
\end{subfigure}%\\
\begin{subfigure}{.50\textwidth}
  \centering
  \includegraphics[width=\linewidth]{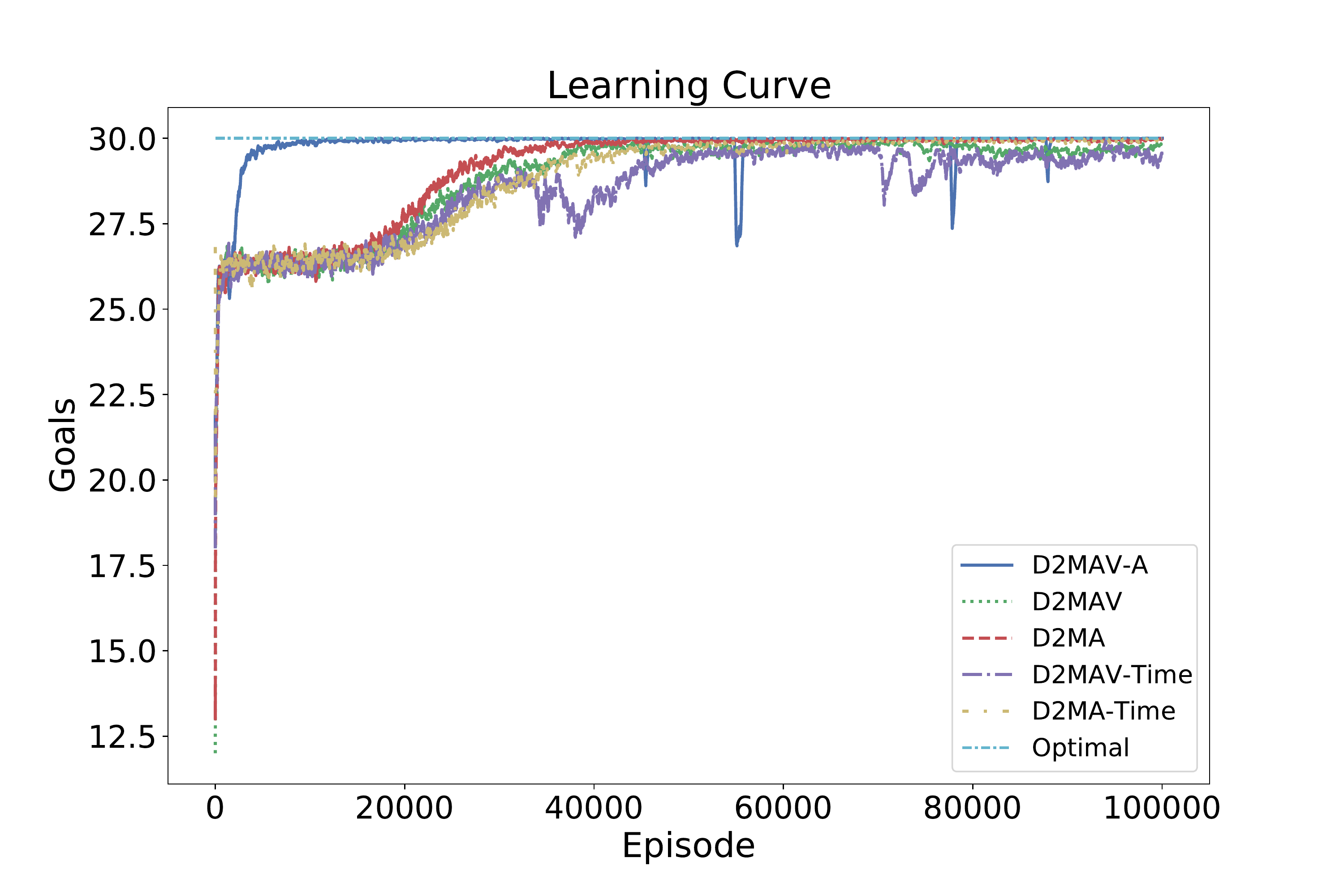}
  \caption{Case Study B}
  \label{fig:lc2}
\end{subfigure}
\begin{center}
\begin{subfigure}{.50\textwidth}
  \centering
  \includegraphics[width=\linewidth]{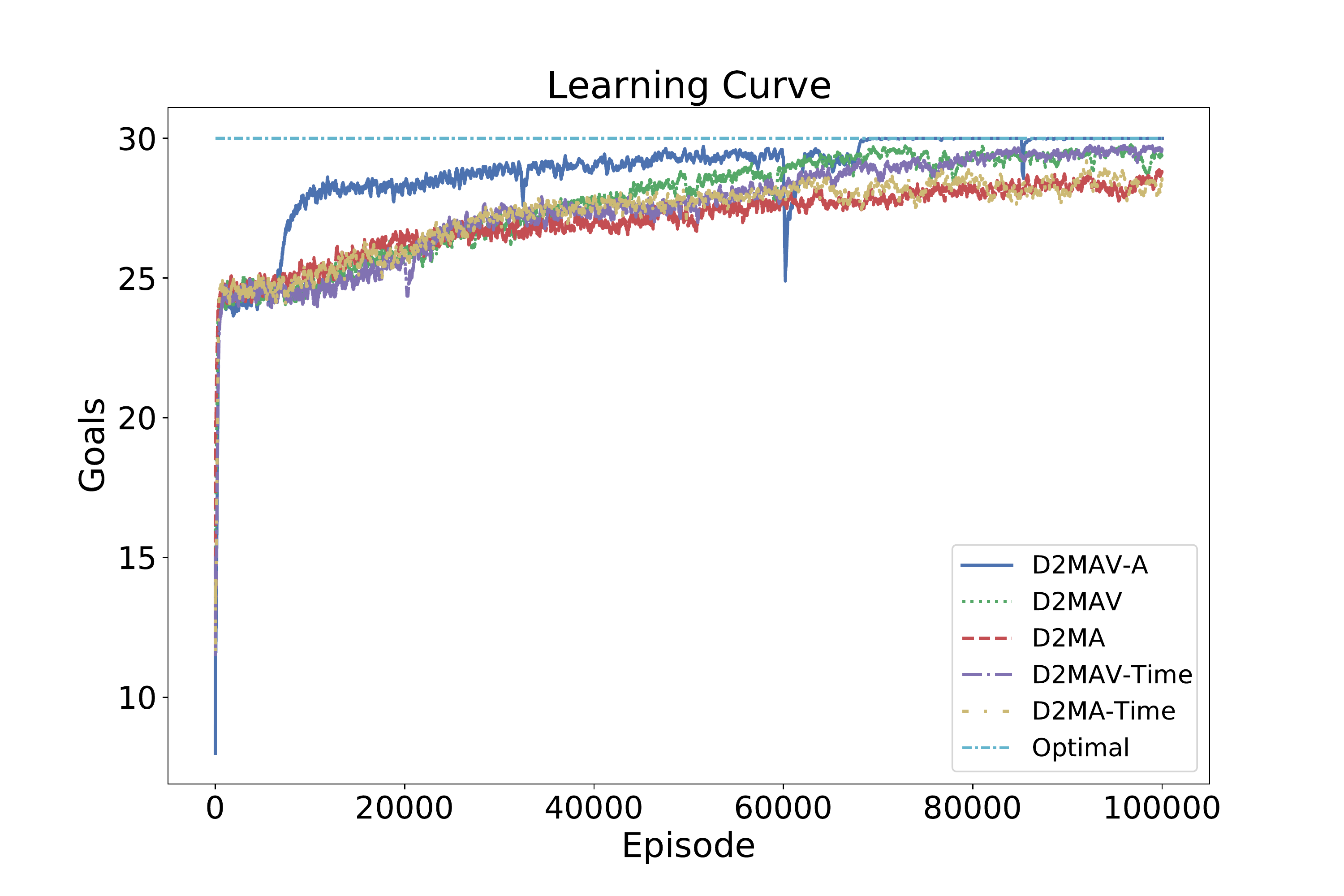}
  \caption{Case Study C}
  \label{fig:lc3}
\end{subfigure}
\end{center}
\caption{Learning Curves smoothed with 150-episode rolling average for clarity.}
\label{learning_curve}
\end{figure*}
\subsection{Case Studies}
In this work, we developed three challenging Case Studies: A, B, and C as shown in Figure~\ref{sector}. In our D2MAV-A framework, the single neural network is implemented on each aircraft as they enter the sector. Each agent is then able to select its own desired speed which greatly increases the complexity of this problem since the agents need to learn how to cooperate in order to maintain safe-separation requirements.

In each case study, aircraft enter the airspace following a uniform distribution over 3 to 6 minutes in 12 second intervals. These are extremely challenging problems to solve because the agents cannot memorize actions, a strategy must be learned in order to achieve optimal performance. The episode terminates when all aircraft have reached their terminal state, so the optimal solution in these case studies is to have all aircraft reach their goal without any loss of separation event. Here a goal is defined as an aircraft exiting the sector without conflict. It is important to note that the difficulty of the case studies is based on the inter-arrival times of the aircraft. The inter-arrival time controls the air traffic density in the airspace\footnote{Shorter inter-arrival times results in higher density airspace since there is less time between aircraft entering the sector on the same route. Likewise, longer inter-arrival times results in a lower density airspace.}, therefore if the agents develop a strategy based on stochastic inter-arrival times there is no limit on the total number of aircraft to send through the sector.

\section{Result Analysis}
In this section, we analyze the performance of the D2MAV-A framework on the case studies and compare it to our baselines. To provide a fair comparison, we allowed all variants to train for 100k episodes to compare the performance. We show that the D2MAV-A framework leads to substantial performance increases in less training time. 

\subsection{Policy Evaluation}
To evaluate the effectiveness of different frameworks, model weights are extracted after training and tested on an additional 200 episodes. During this testing phase, no additional training is allowed, as we are testing the policy on episodes the different frameworks have not seen before. We record the mean, standard deviation, and median for each case study to compare performance. 

As we can see from Table~\ref{policy_perf}, all frameworks obtained near optimal scores on all case studies throughout the 200 episode evaluation phase. The D2MAV-A framework, however leads to substation improvement achieving optimal performance on both case study B and C and near optimal performance on case study A. We can also see that defining the sorting strategy was important for the baselines achieving good performance. This also demonstrates the benefit of the D2MAV-A framework since the sorting strategy does not have to be defined prior, further reducing costly parameter searches. It is also unclear why sorting based on distance is better in some cases and worse in others, questions that are alleviated by the D2MAV-A framework. Given that this is a stochastic environment, we speculate that there could be cases where there is an orientation of aircraft where the 3 nautical mile loss of separation distance can not be achieved, and in such cases we would alert human ATC to resolve this type of conflict.

As mentioned earlier, the D2MAV-A framework can handle any number of aircraft as the density of the airspace is controlled by the inter-arrival times of the aircraft. In Figure~\ref{norm_perf}, we vary the number of aircraft in each case study from 10 to 100. In this evaluation, we do not re-train the model, but use the same model that we tested for 200 episodes in Table~\ref{policy_perf}, only changing the number of aircraft. We allow the model run for 200 episodes, calculate the mean score, and divide the mean score by the total number of aircraft. In this way, the result is normalized such that 1 is the optimal score. We can see from Figure~\ref{norm_perf} that the performance is constant for each case study using the D2MAV-A framework, further showing that our approach is invariant to the number of aircraft sent through the sector.

\subsection{Learning Curve}
The learning curve represents the per-episode returns received during the training phase. This allows one to view the convergence of the frameworks to the optimal policy, along with the speed at which the convergence occurs. We can see from Figure~\ref{learning_curve} that for all case studies: A, B, and C, the D2MAV-A framework is able to quickly learn a good policy, while it takes the other variants much longer to reach comparable performance. In Case Study C, all variants initially follow a similar learning curve, with attention surpassing all other variants by episode 20k.

\subsection{Convergence Speed}
While the learning curve provides a visual overview of the training performance, we quantitatively define convergence as the first time to reach the optimal score over a 150-episode rolling average. With this definition we can record the number of training episodes required by each framework to compare the speed of convergence. 

\begin{table}[htb]
\vskip 0.1cm
\centering
\caption{Number of episodes until convergence. Convergence is defined as the first time to reach the optimal score over a 150 episode rolling average during training.}
\resizebox{6.5cm}{!}{%
\begin{tabular}{llll}
                                    & \multicolumn{3}{c}{Case Study}                                           \\ \cline{2-4} 
\multicolumn{1}{l|}{Framework}      & \multicolumn{1}{c|}{A} & \multicolumn{1}{c|}{B} & \multicolumn{1}{c|}{C} \\ \hline
\multicolumn{1}{l|}{\textbf{D2MAV-A}}      & \multicolumn{1}{l|}{\textbf{3994}}  & \multicolumn{1}{l|}{\textbf{17390}}  & \multicolumn{1}{l|}{\textbf{68912}}  \\
\multicolumn{1}{l|}{D2MAV}           & \multicolumn{1}{l|}{80398}  & \multicolumn{1}{l|}{-}  & \multicolumn{1}{l|}{-}  \\
\multicolumn{1}{l|}{D2MA}      & \multicolumn{1}{l|}{35459}  & \multicolumn{1}{l|}{44278}  & \multicolumn{1}{l|}{-}  \\
\multicolumn{1}{l|}{D2MAV-Time}      & \multicolumn{1}{l|}{43745}  & \multicolumn{1}{l|}{-}  & \multicolumn{1}{l|}{-}  \\
\multicolumn{1}{l|}{D2MA-Time} & \multicolumn{1}{l|}{50233}  & \multicolumn{1}{l|}{79962}  & \multicolumn{1}{l|}{-}  \\ \hline
\end{tabular}

}
\label{converge}
\vskip 0.0cm
\end{table}

In Table~\ref{converge} we compare the number of episodes to convergence for the different frameworks. We can see that for Case Study A and B, D2MAV-A converged in significantly fewer episodes than any of the other variant, with D2MAV-A being the only framework that converged on case study C. This shows that the attention framework can provide a good policy with much fewer computational resources.

\subsection{Wall Clock Time}

As mentioned earlier, D2MAV-A introduces parallel actors to simultaneously collect experience on different instantiations of the same environment. This allows us to make use of modern high-performance computing environments to greatly reduce the training time of the frameworks while also improving performance. It is important to note that great performance increases are available with personal workstations. In this work, a custom workstation was used with an NVIDIA RTX 2080 TI graphics card with an AMD Ryzen Threadripper 2950x CPU (16 cores, 32 threads).

Training for 100k episodes for the D2MAV or D2MA frameworks equates to around 5-8 days of training, but only around 3-5 hours with the D2MAV-A framework. This is due to using parallelized actors in the environment instead of sequentially collecting episodes of experience to update the model.

\subsection{Efficacy of Learned Policy}
\begin{figure*}[tb]
\begin{subfigure}{0.50\textwidth}
  \centering
  \includegraphics[width=\linewidth]{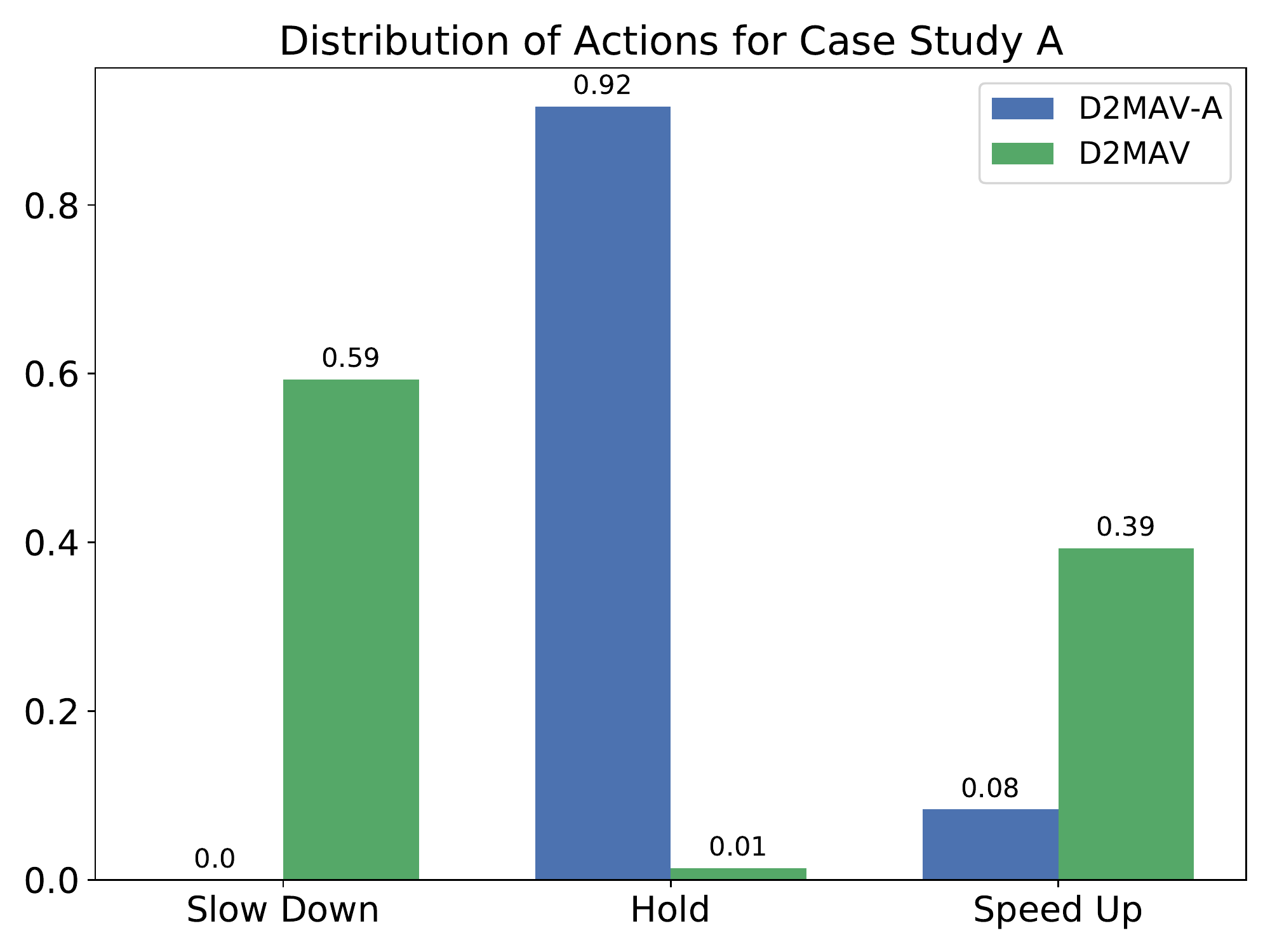}
  \caption{Case Study A}
  \label{fig:ad1}
\end{subfigure}%\\
\begin{subfigure}{.50\textwidth}
  \centering
  \includegraphics[width=\linewidth]{images/action_dist_A.pdf}
  \caption{Case Study B}
  \label{fig:ad2}
\end{subfigure}
\begin{center}
\begin{subfigure}{.50\textwidth}
  \centering
  \includegraphics[width=\linewidth]{images/action_dist_A.pdf}
  \caption{Case Study C}
  \label{fig:ad3}
\end{subfigure}
\end{center}
\caption{Distribution of actions from the learned policy of the D2MAV-A framework compared to D2MAV. We can see that the D2MAV-A framework maximizes the number of holding actions to ensure unnecessary speed changes are avoided.}
\label{policy_efficacy}
\end{figure*}

The D2MAV-A framework introduced a new reward function to achieve two objectives: maintain safe separation between all aircraft and minimize speed changes, where as D2MAV and D2MA did not penalize speed changes. The results in a more efficient policy where speed changes are not constantly advised to the aircraft. We compare the distribution of actions over the evaluation phase for the D2MAV-A framework and D2MAV framework for each case study as shown in Figure~\ref{policy_efficacy}. 

As we can see from Figure~\ref{policy_efficacy}, D2MAV-A results in a significant reduction in speed change actions as compared to the D2MAV framework, without sacrificing performance. This shows that are learned policy is able to maintain safe separation while minimizing the number of speed changes needed.

\subsection{Transfer Learning}

As we can see, the D2MAV-A framework leads to substantial performance increases while also significantly reducing training time. We notice, however that there is still an initial up front training cost on the case studies due to the fact the neural network's weights are initialized with uniform random values. In transfer learning, the idea is to take a policy trained on one environment and use it on a new environment, potentially resulting in better upfront performance and faster convergence speed.

To demonstrate how transfer learning can lead to better performance on new environments we introduced a new case study, case study D that is the combination of all prior case studies; A, B, and C. In case study D, one episode may be the case study A configuration and the next may be case study B. The route configuration is chosen at uniform random at the beginning of each episode. In this case, the neural network must learn a policy that performs well on all of the case studies. We train the D2MAV-A framework on case study D for two configurations: (1) neural network weights initialized with the converged weights from case study C and (2) neural network weights initialized with uniform random values (training from \textit{scratch)}. 

\begin{figure}[t]
\begin{center}
\vskip -0.2cm
\centerline{\includegraphics[width=0.75\columnwidth]{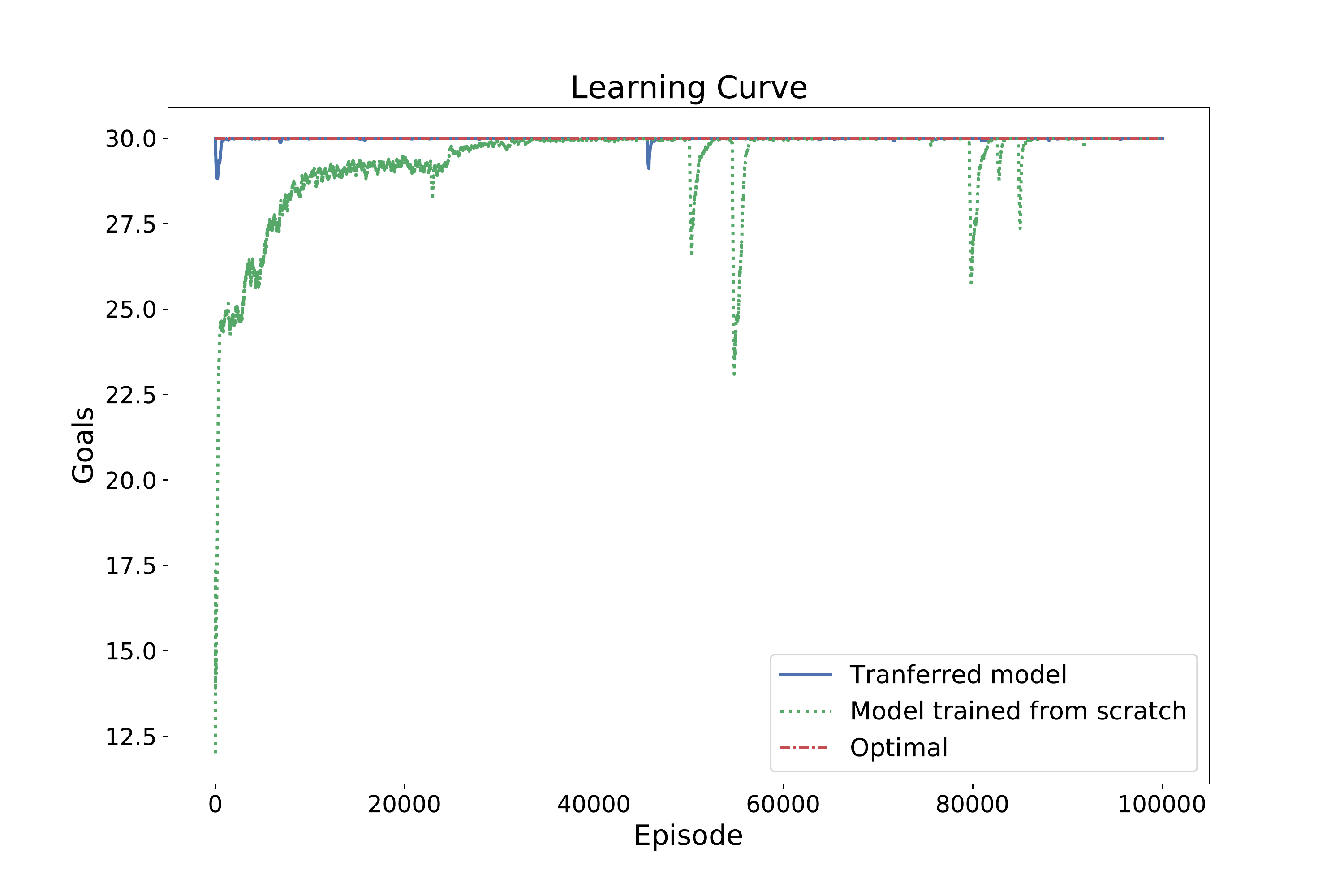}}
\caption{Learning curve for the D2MAV-A framework on case study D for the two variants. The transferred model was initially trained on case study C and then used as initial weights for case study D. The model trained from scratch variant used uniform random weights as if no trained weights were available. Results are smoothed with 150-episode rolling average for clarity.}
\label{LC_transfer}
\end{center}
\vskip -0.2cm
\end{figure}

\begin{table}[htb]
\vskip 0.1cm
\centering
\caption{Number of episodes until convergence for case study D. Convergence is defined as the first time to reach the optimal score over a 150 episode rolling average.}
\resizebox{6.5cm}{!}{%
\begin{tabular}{l|l|}
\cline{2-2}
D2MAV-A Variant & Case Study D \\ \hline
 Transfer         &     \textbf{908}         \\
 Scratch         &      37172        \\ \hline
\end{tabular}}
\label{converge_transfer}
\vskip 0.0cm
\end{table}

As we can see from Table~\ref{converge_transfer}, transfer learning is able to significantly reduce the number of episodes needed to converge from 37172 to 908. 908 episodes only takes around 15-20 minutes with the D2MAV-A framework, resulting in further reduction in the computation needed to solve the case study. The corresponding learning curve is shown in Figure~\ref{LC_transfer} and we can see how in the beginning the policy is able to perform well with a slight decrease in performance in the beginning. We hypothesize that this is due to the model weights experiencing the new environments and adjusting to converge to a policy that works well on all three case studies.

It is important to note that the D2MAV-A framework converged in less episodes on Case Study D as compared to Case Study D. There are two reasons that we believe this occurs: (1) the different case studies introduce different challenges that the agent learns at the same time. In a way this acts as a type of curriculum for the agent to learn simple tasks in the beginning and the progressing to the more complex tasks of case study C. (2) By randomly choosing different case studies with parallel workers, this also acts as a regularizer to prevent overfitting in the neural network, leading to better overall performance.  
%  It is important to note that in the event computational resources are limited, the attention variant not only reduces the number of training episodes until convergence to a good policy, but also eliminates the need to tune additional hyperparameters such as the sorting strategy or number of intruders to consider.

% While encoding the agents based on distance to the ownship through an LSTM is a promising approach, we suspect that important information may be lost for aircraft that are far from the ownship. Case Study C involves an example of where this may occur due to the structure of the routes. Route $R_{2}$ is not structured in the West-East orientation as $R_{1}$ or $R_{3}$. We speculate that to resolve conflicts at $I_{2}$ and $I_{3}$, more planning time is required by the agents, but since the aircraft on $R_{2}$ are further away, their information may be lost in the encoding. One approach to handle this problem is to change the way the intruder information is sorted before processing it through the LSTM. For example, the intruder information could be sorted based on time to the intersection. We leave it to future work to investigate and resolve these types of encounters.

\section{Conclusion}
A novel deep multi-agent reinforcement learning framework is proposed to separate aircraft as a core component in an autonomous separation assurance system in a structured en route sector. We formulate case studies as deep reinforcement learning problems with the actions of speed advisories. We then solve the problem using the D2MAV-A framework, which is shown to be capable of solving complex sequential decision making problems with variable number of agents and uncertainty. By introducing a novel reward function, we demonstrate that improved performance can be obtained with a more efficient policy.

According to our knowledge, the major contribution of this research is that we are the first research group to investigate the feasibility and performance of distributed autonomous aircraft separation with a deep multi-agent reinforcement learning framework that incorporates attention networks to handle variable scalability to enable an automated, safe and efficient en route sector. 

In addition, we also developed the parallelized extension to BlueSky which is reinforcement learning focused. This extension will help drive future research using reinforcement learning techniques in air traffic management.

%%%%%%%%%%%%%%%%%%%%%%%%%%%%%%%%%%%%%%%%%%%%%%%%
\section*{Acknowledgements}
This research is partially funded by the National Science Foundation under Award No. 1718420, NASA Iowa Space Grant under Award No. NNX16AL88H, and the NVIDIA GPU Grant program.

\end{document}